# Sentiment Analysis of Arabic Tweets: Feature Engineering and A Hybrid Approach

**Nora Al-Twairesh, Hend Al-Khalifa, AbdulMalik Alsalman, Yousef Al-Ohali**
College of Computer and Information Sciences
King Saud University
{twairesh, hendk, salman, yousef}@ksu.edu.sa

## Abstract

Sentiment Analysis in Arabic is a challenging task due to the rich morphology of the language. Moreover, the task is further complicated when applied to Twitter data that is known to be highly informal and noisy. In this paper, we develop a hybrid method for sentiment analysis for Arabic tweets for a specific Arabic dialect which is the Saudi Dialect. Several features were engineered and evaluated using a feature backward selection method. Then a hybrid method that combines a corpus-based and lexicon-based method was developed for several classification models (two-way, three-way, four-way). The best F1-score for each of these models was (69.9,61.63,55.07) respectively.

## 1 Introduction

Sentiment Analysis (SA) in Arabic is a challenging task due to the rich morphology of the language. Moreover, the task is further complicated when applied to Twitter data that is known to be highly informal and noisy. The challenges of SA for Arabic were identified in (Al-Twairesh et al., 2014). One of these challenges was the use of Dialectal Arabic (DA). The Arabic language is in a state of diglossia where the formal language used in written form differs radically from the one used in every-day spoken language (Habash, 2010). The formal language is called Modern Standard Arabic (MSA) and the spoken language differs in different Arabic countries producing numerous Arabic dialects. The language used in social media is known to be highly dialectal (Darwish and Magdy, 2014). Previous research on SA of Arabic was merely for MSA, but recently researchers started addressing Dialectal Arabic since people started using dialects in social media text. Dialects differ from MSA phonologically, morphologically and syntactically (Habash, 2010). Moreover, dialects do not have standard orthographies. Most Arabic NLP solutions are designed for MSA and perform poorly on DA (Habash et al., 2012). As (Farghaly and Shaalan, 2009) point out, it is very difficult and almost impossible for one NLP solution to process all the variants of Arabic. As such, an Arabic NLP solution has to specify the Arabic variant it can process beforehand.

Accordingly, in this paper we present a method for sentiment analysis of tweets written in the Saudi Dialect. The Saudi community has witnessed an increased use of Twitter as stated in a study by Semiocast, that the number of twitter users in Saudi Arabia almost doubled in the span of 6 months in 2012 and that Riyadh (the capital city of Saudi Arabia) is now the 10th most active city on Twitter[1]. In a recent study, (Mubarak and Darwish, 2014) used Twitter to collect a multi-dialect corpus of Arabic; a dataset of 175 M Arabic tweets was collected. Then after filtering on tweet user location a subset of 6.5 M tweets was classified according to the tweet's dialectal language, they found that 61% of the tweets were in Saudi dialect followed by 13% Egyptian and 11% Kuwaiti.

The hybrid method proposed in this paper uses a set of features that have been engineered to be dialect-independent and a large corpus of Saudi tweets. It also incorporates a lexicon-based method, hence, the name hybrid. The contributions of this paper are summarized as follows:

---
[1] http://semiocast.com/en/publications/2012_07_30_Twitter_reaches_half_a_billion_accounts_140m_in_the_US Accessed: 14-April-2017

- A set of features for Arabic SA are presented and evaluated using a feature-backward selection method.

- Three classification models for SA of Saudi tweets are developed and compared: two-way (positive and negative), three-way (positive, negative and neutral) and four-way (positive, negative, neutral and mixed)

- The method is also evaluated on an external dataset of a different Arabic dialect to test the proposed features and the feature selection algorithm.

This paper is organized as follows: in section 2, related work is presented. Section 3 presents the feature set and feature-backward algorithm. In section 4 the experiments are presented. The results along with the discussion are presented in section 5. Finally, we conclude the paper in section 6.

## 2 Related Work

Research on SA of Arabic emerged in 2008 with the publication of (Abbasi et al., 2008) which presented a supervised approach to sentiment analysis of both English and Arabic in web forums. Similarly, to English SA, the research on Arabic SA was initially on reviews. Then research on different genres emerged. Surveys on Arabic SA can be found in (Korayem et al., 2012; Al-Twairesh et al., 2014). However, in this section we present a review on Arabic SA for the tweets genre.

Abdul-Mageed et al., (2012) presented the first attempt to SA of Arabic tweets while investigating other genres also. They utilized different features including morphological, lexical and social media features. Also, the effect of the presence of dialectal Arabic in social media text was investigated. Regarding tweets, they found that the use of morphological features was not effective in sentiment classification while using a sentiment lexicon was the more informative feature. Abdulla et al., (2013) compared between a lexicon-based and corpus-based method for tweets written in MSA and the Jordanian dialect. They found that the corpus-based approach using the Support Vector Machine (SVM) classifier and light stemming performed the best. Similarly, (Mourad and Darwish, 2013) used an NB (Naïve-Bayes) classifier along with two lexicons: a translation of an English lexicon and through the use of random graph walk to expand an Arabic lexicon found in previous literature. (Refaee and Rieser, 2014a) engineered a rich set of morphological features, simple syntactic features, stylistic features and semantic features. To obtain the semantic features they used several lexicons: the ArabSenti lexicon (Abdul-Mageed et al., 2011), a translated version of MPQA which was manually corrected, and a manually constructed dialectal lexicon of 484 words extracted from Twitter. The morphological features decreased the performance possibly because the tool used was for MSA and the dataset was highly dialectal. Also, the semantic and stylistic features did not show any significant impact on performance.

A previous study was done on estimating the sentiment of Saudi tweets (Aldayel and Azmi, 2015), the study proposed a hybrid approach, where a lexicon based approach that used a translated lexicon from SentiWordNet and works by counting the number of positive and negative words is used to label the tweets, then a machine learning approach that uses n-grams as a feature in the SVM algorithm takes the labeled tweets as training data to build the classification model. However, the corpus used in this study was very small (1,042) tweets and contained only two classes positive and negative.

In an attempt to study the effect of translation on sentiment, (Salameh et al., 2015) translated English lexicons into Arabic and evaluated their use in Arabic sentiment analysis and translated English tweets into Arabic and evaluated their classification in an Arabic sentiment analysis system. They used an in-house machine translation system and also manual translation to compare between their effects on sentiment. The dataset used constituted of Syrian tweets and a subset of the BBN[2] dataset of weblogs, so the focus of this study was on the Levantine dialect. Although the translated text lost some of the sentiment information as it converted some of the sentiment bearing text into neutral, the system performed very well when compared to current state-of-the art systems of Arabic sentiment analysis.

Although these papers present insights into SA of Arabic, we notice that most of them have used corpora that are relatively small. Moreover, most of the studies do not report what dialect the tweets are written in. In addition, almost all the studies except for (Salameh et al., 2015) have used lexicons that contain MSA words or dialect lexicons that are small also and manually constructed. The language on social media is known to contain slang, non-standard spellings and evolves by time. As such sentiment lexicons that are built from standard dictionaries cannot adequately capture the informal language in social media text. Therefore, in this paper we use Arabic sentiment lexicons that are tweet-specific i.e. generated from tweets.

---

[2] https://catalog.ldc.upenn.edu/LDC2012T09

The methods, size of corpora and performance of these papers are summarized in Table 1.

## 3 Feature Engineering

| Paper | Corpus size | Arabic Variant | Best Result |
|---|---|---|---|
| (Abdulla et al., 2013) | 1000 tweets | Jordanian | 87.2% |
| (Abdul-Mageed et al., 2012) | 3015 tweets | Not specified | 64.37% |
| (Mourad and Darwish, 2013) | 2300 tweets | Not Specified | 72.5% |
| (Refaee and Rieser, 2014) | 4272 tweets | Not Specified | 87.7%:cross validation 38.8%: held out test set |
| (Aldayel and Azmi, 2015) | 1042 tweets | Saudi | 84% |
| (Salameh et al., 2015) | 2000 tweets | Syrian | 79.35% |

Table 1: Summary of work on SA of Arabic tweets

### 3.1 Morphological features

Although several studies on English SA reported improved performance when using POS tags as features, the increase in accuracy is insignificant (Pak and Paroubek, 2010; Kiritchenko et al., 2014). Moreover, the English language has the luxury of a Twitter specific POS tagger: the CMU Twitter NLP tool (Gimpel et al., 2011). The Arabic language does not have such a tool, and most POS taggers for the Arabic language are for MSA with some preliminary work for the Egyptian dialect (Pasha et al., 2014). Also, previous studies on SA of Arabic tweets (Abdul-Mageed et al., 2014; Refaee and Rieser, 2014b) that experimented with using POS tags as features reported that they did not enhance sentiment classification adhering that the POS taggers used were trained on MSA and the newswire domain and were not appropriate for social media text that is highly dialectal. Accordingly, we opted not to use POS tags as features in our classification models.

Other morphological features that have been used in SA of Arabic tweets are: aspect, gender, mood, person, state, and voice. However, (Refaee and Rieser, 2014b) reported that these morphological features actually hurt the performance causing a 21% drop in performance. As such, no morphological features were included in the feature set.

### 3.2 Semantic Features

A sentiment lexicon that contains the prior-polarity of words is used as a feature; i.e. we determine if the tweet contains a positive or negative word according to the sentiment lexicon. Most studies on Twitter SA reported that one of the best features that enhance classification significantly is the use of a sentiment lexicon (Mourad and Darwish, 2013b; Refaee and Rieser, 2014b; Kiritchenko et al., 2014). Therefore, three lexicons were used: the AraSenTi lexicon (Al-Twairesh et al., 2016), the Arabic translations of the lexicons in: MPQA (Wilson et al., 2009) and Bing Liu (Hu and Liu, 2004). The AraSenTi lexicon is a large scale twitter-specific lexicon, i.e. it was extracted from Arabic tweets, hence it captures the peculiarities of Twitter text which motivated us to use it. Other semantic features used are the presence of Arabic negation particles and Arabic contextual valence shifters (intensifiers, diminishers, modal words and contrast words). The modal words can be used to identify the neutral class, since their possible effect is to neutralize sentiment as in (Polanyi and Zaenen, 2006). Also, contrast words can be used to identify the mixed class like for example the word "but" could imply two contradicting sentiments in its before and after clauses. In addition, we used the number of positive and negative words in the tweet according to the AraSenTi lexicon as features. The AraSenTi lexicon also provides the sentiment intensity of each word it contains, we used these numbers to calculate a TweetScore by summing up these numbers for each tweet, this represents the lexicon-based method in the hybrid method we propose.

### 3.3 Stylistic Features

These features check the presence of certain sentiment indicators such as positive and negative emoticons ":)" and ":(". It also includes checking question marks "?" and exclamation marks "!". These features were used by (Refaee and Rieser, 2014b; Kiritchenko et al., 2014) and others. They are included in our feature set also.

### 3.4 Tweet-specific Features

Presence of user mentions (@user), hashtags (#word), URLs, retweet symbol "RT", were used as features in SA of English tweets (Pak and Paroubek, 2010; Go et al., 2009) and in SA of Arabic tweets in (Mourad and Darwish, 2013b). However, they all reported that these features did not have any impact on sentiment classification. Thus, we chose not to include them in the feature set.

Another tweet-specific feature used in research on SA of English tweets is the tweet-length. Kiritchenko et al., (2014) in their state-of the art system on SA of English tweets, reported the use of the number of words in a tweet as a feature. We also include this feature.

### 3.5 Feature Set

The complete set of features used in the classification models to be developed is illustrated in Table 2. A total of 19 features were extracted.

### 3.6 Feature Selection

Feature selection is the process of reducing the feature set used in classification to the subset that yields best performance. The main goal of feature selection is to choose a subset of the feature set, so that the subset can predict the output with accuracy that is comparable to the performance of the complete set but with sufficiently less computational cost. There are different strategies for feature subset selection which include: best-subset selection, forward stepwise selection and backward stepwise selection (Hastie et al., 2011). Best-subset selection performs an exhaustive search on all possible subsets of features until the best classification model that improves classification accuracy is found. Although, as its name implies, it is the best strategy for feature selection; it becomes unfeasible when the number of features is more than 40 since the number of feature subset combinations becomes millions. Alternative strategies are forward and backward stepwise selection. Forward stepwise selection adds features one by one while backward stepwise selection starts with all features in the model and removes them one by one.

Since the impact of features in a learning model is commonly revealed in the combination of these features we chose to start with all the features then do backward selection. This process also allows us to conduct ablation experiments to determine the impact of the features. Each set of features is removed one at a time and the change in performance is observed. The larger the drop in performance, the more useful the removed feature set. However, if the removal of a feature set increases performance, this signals that this feature harms classification and thus should not be included in the final feature set. Although we performed several experiments to assess feature selection algorithms such as Information Gain and Chi Square, we found that our method described above gave the best results. The results of these experiments will be demonstrated in the experiments section.

## 4 Experiments

The proposed hybrid method combines a lexicon-based method and a corpus-based method. The lexicon-based method calculates a TweetScore by utilizing the words' intensities in the AraSenTi lexicon (Al-Twairesh et al., 2016). Then this score is included as a feature in the classification model as we demonstrated in section 3. The corpus based method utilizes a corpus of Saudi tweets (Al-Twairesh et al., 2017) that is presented in the following subsection and an SVM machine learning classifier along with the features presented in Table 2. We used the SVM classifier since it was reported in the majority of studies on SA of tweets to be the best performing classifier (Kiritchenko et al., 2014; Abdul-Mageed et al., 2014). In particular we used the libSVM library (Chang and Lin, 2011) in Weka (Hall et al., 2009) which is an implementation of the SVM algorithm with a linear kernel. The classification was performed on different levels: two-way classification (positive, negative), three-way classification (positive, negative, and neutral) and four-way classification (positive, negative, neutral, and mixed). We present in the following sections the experiments done on each level to reach the best performing classifier.

| Feature | | value |
|---|---|---|
| Semantic | hasPositiveWordAraSenTi<br>hasNegativeWordAraSenTi<br>hasPositiveWordMPQA<br>hasNegaitveWordMPQA<br>hasPositiveWordLiu<br>hasNegaitveWordLiu<br>hasNegation<br>hasIntensifier<br>hasDiminisher<br>hasModalWord<br>hasContrastWord | true, false |
| | PositiveWordCount<br>NegativeWordCount<br>TweetScore | numeric |
| Stylistic | hasQuestionMark<br>hasExclamationMark<br>hasPositiveEmoticon<br>hasNegativeEmoticon | true, false |
| Tweet Specific | tweetLength | numeric |

Table 2: Features used in classification model

The classification is performed using a held out

| Label | Meaning | Label | Meaning |
|---|---|---|---|
| AF | all Features | AL | all-AraSenti lexicon |
| TL | all-TweetLength | WC | all-WordCount |
| QU | all-Question | ML | all-MPQA lexicon |
| EX | all-Exclamation | LL | all-Liu lexicon |
| EM | all-Emoticons | TS | all-TweetScore |
| NE | all-Negation | MW | All-modal words |
| IN | all-Intensifiers | CW | All-contrast words |
| DI | all-Diminishers | | |

Table 4: Labels used in Figures 1,2,3, AF: means all features, TL: means the TweetLength is removed and so on test set. The dataset splits are presented in the following subsection. The reported measures include the F1-score.

### 4.1 Dataset Splits

The corpus used for training and testing consists of tweets written in MSA and the Saudi Dialect(Al-Twairesh et al., 2017). The tweets were manually annotated by three annotators that are Arabic/Saudi native speakers. The conflict between annotators was resolved by majority voting. The corpus contains 17,573 tweets labeled by four labels (positive, negative, neutral, and mixed), to the best of our knowledge this is the largest corpus of Saudi tweets. The dataset splits are illustrated in Table 3. For two-way classification we use only the positive and negative tweets, for three-way classification we use only the positive, negative and neutral tweets, and for four-way classification we use all the datasets. All the reported results are on the test set.

| Class | Training set | Test Set | Total |
|---|---|---|---|
| Positive | 4235 | 722 | 4957 |
| Negative | 5515 | 640 | 6155 |
| Neutral | 4065 | 574 | 4639 |
| Mixed | 1777 | 45 | 1822 |
| Total | 15592 | 1981 | 17573 |

Table 3: Dataset splits: training and testing

### 4.2 Two-way Classification

All the features listed in Table 2 are included in the classification model except for hasModalWord and hasContrastWord which are used to identify neutral and mixed classes; these will be in the following sections. The classifier was run with all the features and the F1-score was calculated, this is illustrated in Figure 1 in the first bar on the left labelled: all features. Then each feature was removed and the F1-score was calculated.

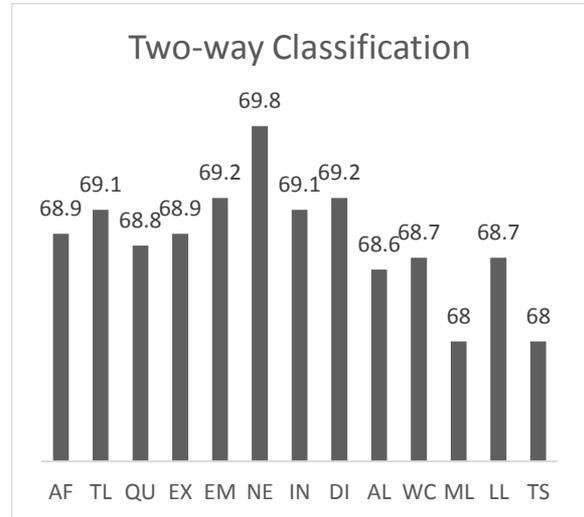

Figure 1: F-score for two-way classification model after removing each feature according to table 4.

### 4.3 Three-way Classification

In this section, we develop the three-way classifier: (positive, negative, and neutral). We perform the same set of experiments as in two-way classification.

We also perform backward feature selection. All the features listed in Table 2 are included in the classification model except for hasContrastWord which is used to identify the mixed class. A total of 18 features were used. The classifier was run with all the features and the F1-score was calculated, this is illustrated in Figure 3, in the first bar on the left labelled: all features. Then each feature set was removed and the F1-score was calculated.

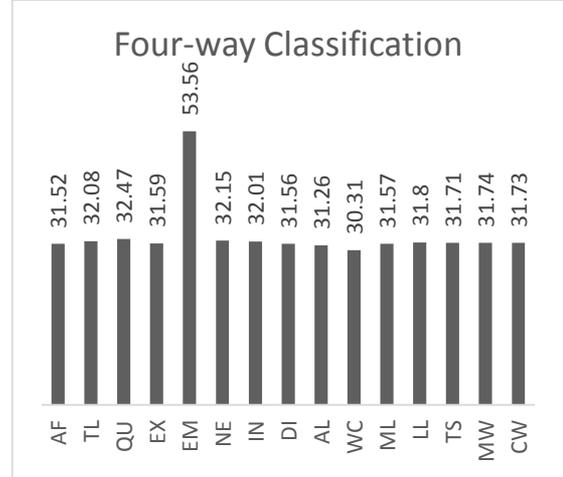

Figure 3: F-score for four-way classification model after removing each feature according to table 4.

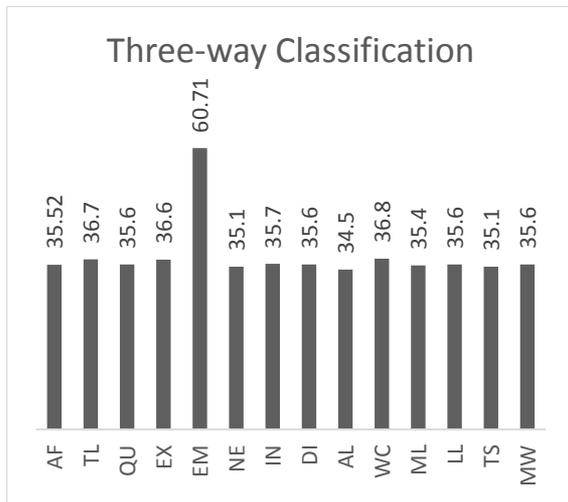

Figure 2: F-score for three-way classification model after removing each feature according to table 4.

### 4.4 Four-way Classification

In this section we develop the four-way classifier: (positive, negative, neutral, and mixed). We perform the same set of experiments as in two-way and three-way classification.

We also perform backward feature selection. All the features listed in Table 2 are included in the classification model. A total of 19 features were used. The classifier was run with all the features and the F1-score was calculated, this is illustrated in Figure 2, in the first bar on the left labelled: all features. Then each feature was removed and the F1-score was calculated.

## 5 Results and Discussion

### 5.1 Two-way Classification

We observe from Figure 1 that the removal of the features: tweetLength, hasPositiveEmoticon, hasNegativeEmoticon, hasNegation, hasIntensifier, and hasDiminisher, caused an increase in F1-score, meaning that these features harm the performance and should be removed. So, accordingly, we removed these features and recalculated the F1-score and found that it increased to **69.5**. However, this is less than the F1-score of removing the hasNegation feature alone which was 69.8. So using the feature set that does not contain hasNegation we also started removing the feature sets one by one and check the F1-score. The best F1-score reached was **69.9** through removing hasNegation and hasDiminisher. A possible reason for the negative impact of hasNegation on the classification is that the Tweet Score already incorporates knowledge on negation since the negation particles have a score in the AraSenTi-lexicon. Also this feature is set to true by checking the presence of a negation particle in the tweet. In some cases, the negation particle does not impact sentiment i.e. it is not used to negate a sentiment word and does not change its polarity like in the following example from the dataset:

أمس بدأت في قراءة " تثريب ". . لسلطان الموسى ,
أسلوبه رائع ما دريت عن نفسي إلا وأنا قاري 170
صفحة و الليلة بخلصه إن شاء الله.

Translation: Yesterday I started reading "Tathreeb" by Sultan Al-Mousa his style is fabulous I didn't know about myself until I had read 170 pages and tonight I will finish it God willing.

This special case where the negation particle does not impact sentiment was found a lot in the dataset.

### 5.2 Three-way Classification

We observe from Figure 3 that the removal of the hasPositiveEmoticon and hasNegativeEmoticon feature set caused a substantial increase in performance (+25%). Other features that also increased performance when removed are: tweetLength, hasQuestionMark, hasExclamationMark, hasModal. hasIntensifier, hasDiminisher, PositiveWordCount, NegativeWordCount, hasPositiveWordLiu and hasNegativeWordLiu. These were all removed. After removing these features the classification model reached an F1-score of **61.5**.

### 5.3 Four-way Classification

We observe from Figure 2 that the removal of the hasPositiveEmoticon and hasNegativeEmoticon feature set had the highest impact: it caused a +23% increase in performance; this is similar to what happened in three-way classification. This could imply the noisy effect of emoticons on sentiment although it was expected to help identify sentiment. The other features that also increased performance when removed were: tweetLength, hasQuestionMark, hasExclamationMark, hasNegation, hasIntensifier, hasDiminisher, hasContrastWord, hasModalWord, hasPositiveWordMPQA, hasNegativeWordMPQA, hasPositiveWordLiu, hasNegativeWordLiu and TweetScore. These features were all removed and the F1-score was calculated for the classifier with the reduced feature set and it was **55.07%** which indicates that four-way classification does not benefit much from the Tweet Score. The reason could be in that the lexicon-based method actually does not use the Tweet Score to identify the mixed class but merely depends on the presence of contrast words to identify the mixed class i.e. the Tweet Score does not incorporate any knowledge for identifying the mixed class. Moreover, we notice that the number of tweets in the corpus for the mixed class is very low compared to the other classes.

### 5.4 Feature Set Impact on Classification

Extensive series of experiments were performed and a general observation was that with each change in one parameter or setting, a new reduced feature set was produced. Interaction between features is different and unpredictable. So we review here the impact of each feature set on all classification models developed in this paper. We also note that statistical tests were performed using the Pearson Chi-Square test which is a statistical test that is used to measure the goodness of fit for the models as following the work of (Taboada et al., 2011). We test if the developed models are significantly different than the human-predicted labels i.e. the actual labels. at a confidence interval of 95% ($p<0.05$) and we found that all the developed classification models were statistically significant.

Emoticons had a negative impact on almost all classification models except for two-way classification, although this feature is widely used as an indicator of sentiment. We had expected such a result when we were manually inspecting the tweets to construct the corpus as we observed that the presence of emoticons does not always infer sentiment. In line with this result, (Abdul-Mageed, 2015) assessed the correlation of emoticons with sentiment and found that positive emoticons do not always express positive sentiment while negative emoticons are usually correlated with negative sentiment. A similar finding was proposed in (Dresner and Herring, 2014) also, where they found that emoticons express the intention of the user not the sentiment.

Tweet length, which was the only tweet-specific feature that was used, also didn't have an impact on improving classification performance. The impact of the presence of question marks and exclamation marks varied but was evident more in two-way classification. However, the impact of contextual valence shifters as features was not consistent across the classification models and a clear impact cannot be deduced.

The AraSenTi lexicon contributed to two feature sets: (hasPositiveWordAraSenti, hasNegativeWordAraSenti) and (PositiveWordCount, NegativeWordCount). These feature sets are the only ones that are present in the best feature set of all three classification models. Although, in the two-way classification hybrid method the drop in performance when removing this feature set was small, this discrepancy could be due to the knowledge of this feature being represented already in the Tweet Score feature. This same conjecture could explain also why the word-count feature was not in the best feature set for the three-way classification. As for the features (hasPositiveWordAraSenti, hasNegativeWordAraSenti) feature set, it is evident in all the classification models. This exhibits the significance of having a lexicon that is extracted from Twitter on sentiment classification.

### 5.5 External Comparison/Evaluation

When comparing the performance of the developed classifiers to the state-of-the-art in SA of English tweets (Mohammad et al., 2013; Zhu et al., 2014), which was the winning system in SemEval 2013 and 2014 for the task of Sentiment Analysis of Twitter, their best F1-score was 70.45. As for the winning system in SemEval 2015 (Büchner and Stein, 2015), the best F1-score was 64.84.

SA of Arabic tweets lacks a benchmark that we can compare to. However, the most cited papers on SA of Arabic tweets are those of (Abdul-Mageed et al., 2014; Mourad and Darwish, 2013b) their F1-score was 64.37 and 72 respectively. Details of all these papers were illustrated in the related work in section2. The performance of our classifiers is quite comparable to these results.

Although a proper benchmark for Arabic SA is not available, there was an attempt by (Nabil et al., 2015) in which they constructed a dataset of 10,000 tweets and performed four-way sentiment classification using only n-gram features in an effort to provide a benchmark for Arabic sentiment classification, they call this dataset ASTD. The best F1-score was 62.6. However, the dataset was collected from Egyptian twitter accounts and trending hashtags and thus the tweets are in the Egyptian dialect. As we stated before that according to the literature Arabic dialects differ substantially and a solution for one dialect will not work for another. We demonstrate this through the following experiment where we apply our four-way classification model on this dataset. In this experiment, the training set contains tweets of the Saudi dialect and the test set contains tweets from the Egyptian dialect. The F1-score was 53.68, as expected the performance degraded due to the difference in the Arabic dialect between the training and test sets.

Nonetheless, we attempt to validate the feature sets we engineered including the AraSenTi lexicon on the ASTD dataset without using our corpus as the training set but by dividing the ASTD dataset into training and test sets using the same splits we had used on our corpus 89-11 % for training and testing respectively. We performed feature backward selection in the same way mentioned before. The best F1-score reached after feature backward selection was 65.8, this is a +3.2% improvement on the performance of the benchmark which was 62.6. This exhibits the significance of the engineered features in this paper and the feature backward selection algorithm when applied on an external dataset.

## 6 Conclusion

In this paper, we presented three hybrid sentiment analysis classifiers for Arabic tweets. The classifiers work on different levels of classification: two-way classification (positive and negative), three-way classification (positive, negative, and neutral) and four-way classification (positive, negative, neutral, and mixed). The approach was to incorporate the knowledge extracted from the lexicon-based method as features into the corpus-based method to develop the hybrid method. First, a set of features were extracted from the data then a backward selection algorithm was proposed to perform feature selection in order to reach the best classification performance.

The impact of the proposed feature sets on all the developed classification models was investigated. The feature sets that were present in all the best feature sets of all three classification models are the ones extracted from the AraSenTi lexicon. The impact of other feature sets varied, between those that had a negative impact on most classification models such as the emoticons features and the tweet length feature and those that didn't exhibit a clear impact on sentiment classification.

### Acknowledgments

This Project was funded by the National Plan for Science, Technology and Innovation (MAARIFAH), King Abdulaziz City for Science and Technology, Kingdom of Saudi Arabia, Award Number (GSP-36-332.

### References

Ahmed Abbasi, Hsinchun Chen, and Arab Salem. 2008. Sentiment analysis in multiple languages: Feature selection for opinion classification in Web forums. *ACM Transactions on Information Systems (TOIS)*, 26(3):12.

Nawaf A. Abdulla, Nizar A. Ahmed, Mohammed A. Shehab, and Mahmoud Al-Ayyoub. 2013. Arabic sentiment analysis: Lexicon-based and corpus-based. In *Applied Electrical Engineering and Computing Technologies (AEECT),* pages 1–6. IEEE.

Muhammad Abdul-Mageed. 2015. *Subjectivity and sentiment analysis of Arabic as a morophologically-rich language*. Ph.D. thesis, INDIANA UNIVERSITY.

Muhammad Abdul-Mageed, Mona Diab, and Sandra Kübler. 2014. SAMAR: Subjectivity and sentiment analysis for Arabic social media. *Computer Speech & Language*, 28(1):20–37, January.

Muhammad Abdul-Mageed, Mona T. Diab, and Mohammed Korayem. 2011. Subjectivity and sentiment analysis of modern standard arabic. In *Proceedings of the 49th Annual Meeting of the Association for Computational Linguistics: Human Language Technologies: short papers-Volume 2*, pages 587–591. Association for Computational Linguistics.

Muhammad Abdul-Mageed, Sandra Kübler, and Mona Diab. 2012. Samar: A system for subjectivity and sentiment analysis of arabic social media. In *Proceedings of the 3rd Workshop in Computational Approaches to Subjectivity and Sentiment Analysis*, pages 19–28. Association for Computational Linguistics.

Haifa K Aldayel and Aqil M Azmi. 2015. Arabic tweets sentiment analysis–a hybrid scheme. *Journal of Information Science*. in press.

Nora Al-Twairesh, Hend Al-Khalifa, and AbdulMalik Al-Salman. 2016. AraSenTi: Large-Scale Twitter-Specific Arabic Sentiment Lexicons. In *Proceedings of the 54th Annual Meeting on Association for Computational Linguistics*, Berlin, Germany. Association for Computational Linguistics.

Nora Al-Twairesh, Hend Al-Khalifa, AbdulMalik Al-Salman, and Yousef Al-Ohali. 2017. AraSenTi-Tweet: A Corpus for Arabic Sentiment Analysis of Saudi Tweets. In *Procedia Computer Science*, volume 117, pages 63–72, Dubai. Elsevier.

Nora Al-Twairesh, Hend S. Al-Khalifa, and AbdulMalik Al-Salman. 2014. Subjectivity and Sentiment Analysis of Arabic : Trends and Challenges. In *The 11th ACS/IEEE International Conference on Computer Systems and Applications (AICCSA' 2014)*, Doha, Qatar.

Matthias Hagen Martin Potthast Michel Büchner and Benno Stein. 2015. Webis: An Ensemble for Twitter Sentiment Detection. In *The 9th International Workshop on Semantic Evaluation*, page 582, Denver, Colorado, USA, June.

Chih-Chung Chang and Chih-Jen Lin. 2011. LIBSVM: a library for support vector machines. *ACM Transactions on Intelligent Systems and Technology (TIST)*, 2(3):27.

Kareem Darwish and Walid Magdy. 2014. Arabic Information Retrieval. *Foundations and Trends in Information Retrieval*, 7(4):239–342.

Eli Dresner and Susan C Herring. 2014. Emoticons and illocutionary force. In *Perspectives on Theory of Controversies and the Ethics of Communication*, pages 81–90. Springer.

Ali Farghaly and Khaled Shaalan. 2009. Arabic natural language processing: Challenges and solutions. *ACM Transactions on Asian Language Information Processing (TALIP)*, 8(4):14.

Kevin Gimpel, Nathan Schneider, Brendan O'Connor, Dipanjan Das, Daniel Mills, Jacob Eisenstein, Michael Heilman, Dani Yogatama, Jeffrey Flanigan, and Noah A Smith. 2011. Part-of-speech tagging for twitter: Annotation, features, and experiments. In *Proceedings of the 49th Annual Meeting of the Association for Computational Linguistics: Human Language Technologies: short papers-Volume 2*, pages 42–47. Association for Computational Linguistics.

Alec Go, Richa Bhayani, and Lei Huang. 2009. Twitter sentiment classification using distant supervision. *CS224N Project Report, Stanford*:1–12.

Nizar Habash, Ramy Eskander, and Abdelati Hawwari. 2012. A morphological analyzer for Egyptian Arabic. In *Proceedings of the Twelfth Meeting of the Special Interest Group on Computational Morphology and Phonology*, pages 1–9. Association for Computational Linguistics.

Nizar Y Habash. 2010. Introduction to Arabic natural language processing. *Synthesis Lectures on Human Language Technologies*, 3(1):1–187.

Mark Hall, Eibe Frank, Geoffrey Holmes, Bernhard Pfahringer, Peter Reutemann, and Ian H Witten. 2009. The WEKA data mining software: an update. *ACM SIGKDD explorations newsletter*, 11(1):10–18.

Trevor J. Hastie, Robert John Tibshirani, and Jerome H Friedman. 2011. *The elements of statistical learning: data mining, inference, and prediction*. Springer, Second edition.

Minqing Hu and Bing Liu. 2004. Mining and summarizing customer reviews. In *Proceedings of the tenth*

*ACM SIGKDD international conference on Knowledge discovery and data mining*, pages 168–177. ACM.

Svetlana Kiritchenko, Xiaodan Zhu, and Saif M Mohammad. 2014. Sentiment analysis of short informal texts. *Journal of Artificial Intelligence Research*, 50:723–762.

Mohammed Korayem, David Crandall, and Muhammad Abdul-Mageed. 2012. Subjectivity and sentiment analysis of arabic: A survey. In *Advanced Machine Learning Technologies and Applications*, pages 128–139. Springer.

Saif M Mohammad, Svetlana Kiritchenko, and Xiaodan Zhu. 2013. NRC-Canada: Building the State-of-the-Art in Sentiment Analysis of Tweets. In *Proceedings of the 8th International Workshop on Semantic Evaluation (SemEval 2013)*, page 321, Atlanta, Georgia, USA.

Ahmed Mourad and Kareem Darwish. 2013a. Subjectivity and Sentiment Analysis of Modern Standard Arabic and Arabic Microblogs. In *Proceedings of the 4th workshop on computational approaches to subjectivity, sentiment and social media analysis.WASSA 2013*, pages 55–64.

Ahmed Mourad and Kareem Darwish. 2013b. Subjectivity and Sentiment Analysis of Modern Standard Arabic and Arabic Microblogs. *WASSA 2013*:55.

Hamdy Mubarak and Kareem Darwish. 2014. Using Twitter to collect a multi-dialectal corpus of Arabic. *ANLP 2014*:1.

Mahmoud Nabil, Mohamed Aly, and Amir F Atiya. 2015. ASTD: Arabic Sentiment Tweets Dataset. In *Proceedings of the 2015 Conference on Empirical Methods in Natural Language Processing*, pages 2515–2519.

Alexander Pak and Patrick Paroubek. 2010. Twitter as a Corpus for Sentiment Analysis and Opinion Mining. In *Proceedings of the Sixth International Conference on Language Resources and Evaluation (LREC 2010)*, Valleta,Malta. European Language Resources Association (ELRA).

Arfath Pasha, Mohamed Al-Badrashiny, Ahmed El Kholy, Ramy Eskander, Mona Diab, Nizar Habash, Manoj Pooleery, Owen Rambow, and Ryan Roth. 2014. Madamira: A fast, comprehensive tool for morphological analysis and disambiguation of arabic. In *Proceedings of the 9th International Conference on Language Resources and Evaluation, LREC 2014*, Reykjavik, Iceland. European Language Resources Association (ELRA).

Livia Polanyi and Annie Zaenen. 2006. Contextual valence shifters. In *Computing attitude and affect in text: Theory and applications*, pages 1–10. Springer.

Eshrag Refaee and Verena Rieser. 2014a. Subjectivity and Sentiment Analysis of Arabic Twitter Feeds with Limited Resources. In *Workshop on Free/Open-Source Arabic Corpora and Corpora Processing Tools Workshop Programme*, page 16.

Eshrag Refaee and Verena Rieser. 2014b. Subjectivity and Sentiment Analysis of Arabic Twitter Feeds with Limited Resources. In *Workshop on Free/Open-Source Arabic Corpora and Corpora Processing Tools Workshop Programme*, page 16.

Mohammad Salameh, Saif M Mohammad, and Svetlana Kiritchenko. 2015. Sentiment after translation: A case-study on arabic social media posts. In *Proceedings of the 2015 Conference of the North American Chapter of the Association for Computational Linguistics: Human Language Technologies*, pages 767–777.

Maite Taboada, Julian Brooke, Milan Tofiloski, Kimberly Voll, and Manfred Stede. 2011. Lexicon-based methods for sentiment analysis. *Computational linguistics*, 37(2):267–307.

Theresa Wilson, Janyce Wiebe, and Paul Hoffmann. 2009. Recognizing contextual polarity: An exploration of features for phrase-level sentiment analysis. *Computational linguistics*, 35(3):399–433.

Xiaodan Zhu, Svetlana Kiritchenko, and Saif M Mohammad. 2014. Nrc-canada-2014: Recent improvements in the sentiment analysis of tweets. In *Proceedings of the 8th International Workshop on Semantic Evaluation (SemEval 2014)*, pages 443–447.